\documentclass[pmlr]{jmlr}

\usepackage{longtable}
\usepackage{multirow}
\usepackage{caption}
\usepackage{hyperref}
\usepackage{natbib} 

\title{Facts Do Care About Your Language: Assessing Answer Quality of Multilingual LLMs}

\author{
\begin{center}
Shmuel Berman\thanks{These authors contributed equally. The order was determined by a coin toss.}, 
Yuval Kansal\footnotemark[1], 
Lydia Liu \\
Princeton University, USA \\
\texttt{\{shmuel, yuvalkansal, ltliu\}@princeton.edu}
\end{center}
}

\begin{document}

\maketitle
\thispagestyle{empty}
\begin{abstract}
Factuality is a necessary precursor to useful educational tools. As adoption of Large Language Models (LLMs) in education continues of grow, ensuring correctness in all settings is paramount. Despite their strong English capabilities, LLM performance in other languages is largely untested. In this work, we evaluate the correctness of the Llama3.1 family of models in answering factual questions appropriate for middle and high school students. We demonstrate that LLMs not only provide extraneous and less truthful information, but also exacerbate existing biases against rare languages.

\end{abstract}

\begin{keywords}
AI in Education, LLMs, Question Answering, Factuality
\end{keywords}

\section*{Introduction}
\label{sec:intro}
LLMs have shown exciting human-like capabilities in tasks such as text generation and question answering. Tools such as ChatGPT and LLama are being used in educational learning contexts and are employed directly by students without expert guidance. As they become ubiquitous learning tools, it is vital to ensure they reduce rather than reinforce existing inequities. Equitable access to education is essential for the betterment of society. The inconsistencies in LLMs for providing information to simple questions across multiple languages are concerning. In this study, we assess the factuality and consistency of LLM responses to middle and high school level questions in non-English languages.

About 55\% of the content on the internet is in English \cite{wiki:Languages_used_on_the_Internet}.
All other languages  make up a fraction of the content, with some accounting for only ~0.1-0.2\% of the internet websites. It has been shown that a LLM's effectiveness is heavily dependent on the distribution and quality of training data \cite{xie2024quantifying}. LLMs trained on poorly distributed data can undergo model collapse, gradually degenerate and produce increasingly homogeneous outputs \cite{du2024diversity}.

Recent studies have shown that ``dominant" languages exhibit superior performance due to their substantial representation in training data, while low-resource languages demonstrate significantly degraded capabilities \cite{xie2024quantifying,ponti2019modeling}.  Insufficient training data in a language can result in the model losing valuable information about that language \cite{du2024diversity,lankford2024llm}. While prior work has evaluated language fluency and translation quality of LLMs 
 \cite{xie2024quantifying,ponti2019modeling}, our aim is to measure their factuality in different languages, especially in educational settings.


We present three initial contributions in evaluating LLM factuality: (1) A benchmark of 54 factual questions and corresponding keyword-level answers spanning various subjects  (2) A manual evaluation (IRB approved) of translated answer quality conducted by bilingual individuals (3) A statistical evaluation of several LLMs' responses across languages which shows poorer performance across several metrics in non-English languages.

\section*{Related Work}
Prior research has shown that LLMs tend to perform better in  English and French compared to  Bengali, Georgian, Nepali, and Maithili \cite{dong_evaluating_2024}. The paper highlights safety alignment issues, the discrepancy in multilingual research, and proposes a new safety multilingual benchmark.``Factcheck-Bench" introduces a similar English-only dataset to evaluate LLM factual accuracy and proposes a pipeline to get better results \cite{wang_factcheck-bench_2024} These datasets limit their evaluation to factuality and do not take into account extraneous information.

Researchers at Georgia Tech found that LLMs gave richer and higher-quality responses in English than in other languages \cite{jin_better_2023}. Wang et al. showed that ChatGPT performs much better at translating English to German than to Chinese \cite{wang_document-level_2023}. Li et al. showed that there is a direct relationship between a specific LLM’s performance in a language and the percentage of data belonging that language in the training corpus \cite{li_quantifying_2024}. Zhu et al. showcased inconsistency across non-English languages and successfully boosted LLM performance in instruction following and translation through multilingual instruction tuning 
 \cite{zhu_extrapolating_2023}. These studies quantify translation quality and contribute valuable benchmarks but do not discuss the factual information content.

\vspace{-5mm}
\section*{Methodology}
To assess LLM factuality, we compiled a list of 54 questions across different subjects covering middle and high school curriculum. Each question was designed to be answerable in a single sentence and is accompanied by manually curated target words essential for a correct response. For example, for the question ``What is photosynthesis?'', target words included ``sunlight'',``carbon dioxide'', and ``glucose''.

These questions were presented to Llama 3.1 (8B \& 70B). The evaluation was conducted in 12 languages—Japanese, Arabic, French, Chinese, Persian, Hebrew, Hindi, Nepali, Haitian Creole, Tulu, Māori, and English. All responses were machine-translated back into English using the Google Translate API for analysis. While machine translation introduces potential limitations, prior research has demonstrated that Google Translate achieves adequate accuracy for tasks requiring factual content. Taira et al. assessed Google Translate's performance and reported mean Likert scale adequacy scores exceeding 4 across languages \cite{taira2021pragmatic}. Our study focuses on factual accuracy, not fluency. Google Translate is sufficient for this purpose given that literal translations often suffice to convey essential information.

For the manual analysis, bilingual evaluators assessed each response on a multidimensional rubric: counting the presence
of each target keyword, factual accuracy (1-3 scale), and extraneous information (1-5 scale). 
For our automated analysis, we search for the presence of target words in both the original language and in the response translated to English, perform keyword extraction, and measure per-language response properties such as keyword repetition. We use KeyBERT \cite{grootendorst2020keybert} to identify semantically important terms and extract them as keywords, and include metrics for the number of keywords and how many times they repeat. Our question bank and code can be found \href{https://github.com/shmublu/multilingual-factuality/tree/main}{here}.

\section*{Results}
We conducted a manual evaluation of Llama-3.1-8B and 70B responses on English, Hindi, Hebrew, and present our averaged findings in the table below:

\begin{table}[h]
\centering
\resizebox{\textwidth}{!}{
\begin{tabular}{lcrrrr}
\hline
Language & Model & \# Speakers (millions) & Keyword Coverage (\%) & Incorrectness \footnotemark[1] & Extraneous Score\footnotemark[2] \\
\hline
\multirow{2}{*}{Hebrew}  & LLaMA 8B  & 9   & 27.28  & 1.93  & 3.48  \\
                         & LLaMA 70B & 9   & 69.25  & 1.44  & 2.11  \\
\hline
\multirow{1}{*}{Farsi}   & LLaMA 70B & 110 & 87.36  & 1.06  & 1.13  \\
\hline
\multirow{2}{*}{Hindi}   & LLaMA 8B  & 345 & 51.17  & 1.50  & 2.13  \\
                         & LLaMA 70B & 345 & 77.25  & 1.13  & 1.30  \\
\hline
\multirow{2}{*}{English} & LLaMA 8B  & 1500 & 94.44 & 1.04  & 1.35  \\
                         & LLaMA 70B & 1500 & 97.48 & 1.09  & 1.26  \\
\hline
\end{tabular}
}
\captionsetup{font=small, labelfont=it}
\caption{\textit{Manual Analysis Results for LLaMA 8B and LLaMA 70B. Keyword Coverage (higher is better), Incorrectness Score\protect \footnotemark[1], and Extraneous Score\protect \footnotemark[2] (lower is better).}}
\label{tab:combined_results}
\end{table}

\footnotetext[1]{\textbf{Incorrectness Score}: Scale 1-3 — 1 if no incorrect information, 2 if some incorrect information, and 3 if clearly incorrect information.}
\footnotetext[2]{\textbf{Extraneous Score}: Scale 1-5 — 1 if no extraneous information, 5 if most of the answer is unnecessary.}

To analyze the relationship between language representation and response quality, we present the Spearman Correlation coefficients for target word detection correlating the number of speakers of the language of each response with its features. We determine there is a statistically significant correlation between the presence of our hand-picked target words. For the larger model, there is also a strong correlation between the raw number of KeyBERT keywords, which are question-agnostic. We conjecture this is because smaller models give terser responses across all languages, but leave formal analysis to future work.
\begin{table}[h]
\centering
\begin{tabular}{lcrr}
\hline
Feature & Model & Correlation & p-Value \\
\hline
\multirow{2}{*}{Original Target Words Count} & LLaMA 8B  & 0.426310 & 1.535e-27  \\
                                             & LLaMA 70B & 0.463306 & 6.1005e-33 \\
\hline
\multirow{2}{*}{Translated Target Words Count} & LLaMA 8B  & 0.372825 & 5.841e-21  \\
                                               & LLaMA 70B & 0.400362 & 2.8294e-24 \\
\hline
\multirow{2}{*}{KeyBERT Keywords Count} & LLaMA 8B  & 0.074542 & 6.993e-02  \\
                                        & LLaMA 70B & 0.348306 & 2.2022e-18 \\
\hline
\multirow{2}{*}{KeyBERT Keyword Repetition} & LLaMA 8B  & 0.063026 & 1.256e-01  \\
                                            & LLaMA 70B & 0.345928 & 3.8637e-18 \\
\hline
\multirow{1}{*}{Word Repetition} & LLaMA 70B & 0.372388 & 5.6222e-21 \\
\hline
\multirow{1}{*}{English Answer Length} & LLaMA 70B & 0.423705 & 2.8087e-27 \\
\hline
\end{tabular}
\captionsetup{font=small, labelfont=it} 
\caption{\textit{Correlation Analysis Results between \# of Speakers and Features for LLaMA 8B and 70B.}}
\label{tab:combined_correlation_results}
\end{table}
\vspace{-5mm}
\section*{Conclusion}
Our experiments confirm our hypothesis that LLMs are less factual in languages that have fewer speakers and are less-represented in available training material. Future work will focus on extending this study to a broader range of LLMs and refining the analysis by utilizing professional human translators rather than machine translation. By identifying potential linguistic biases embedded within the training data of these models, we aim to enable researchers to create more equitable language models.

\bibliography{references}

\appendix
\section*{Appendix}
\begin{figure}[htbp]
    \centering
    \includegraphics[width=0.6\textwidth]{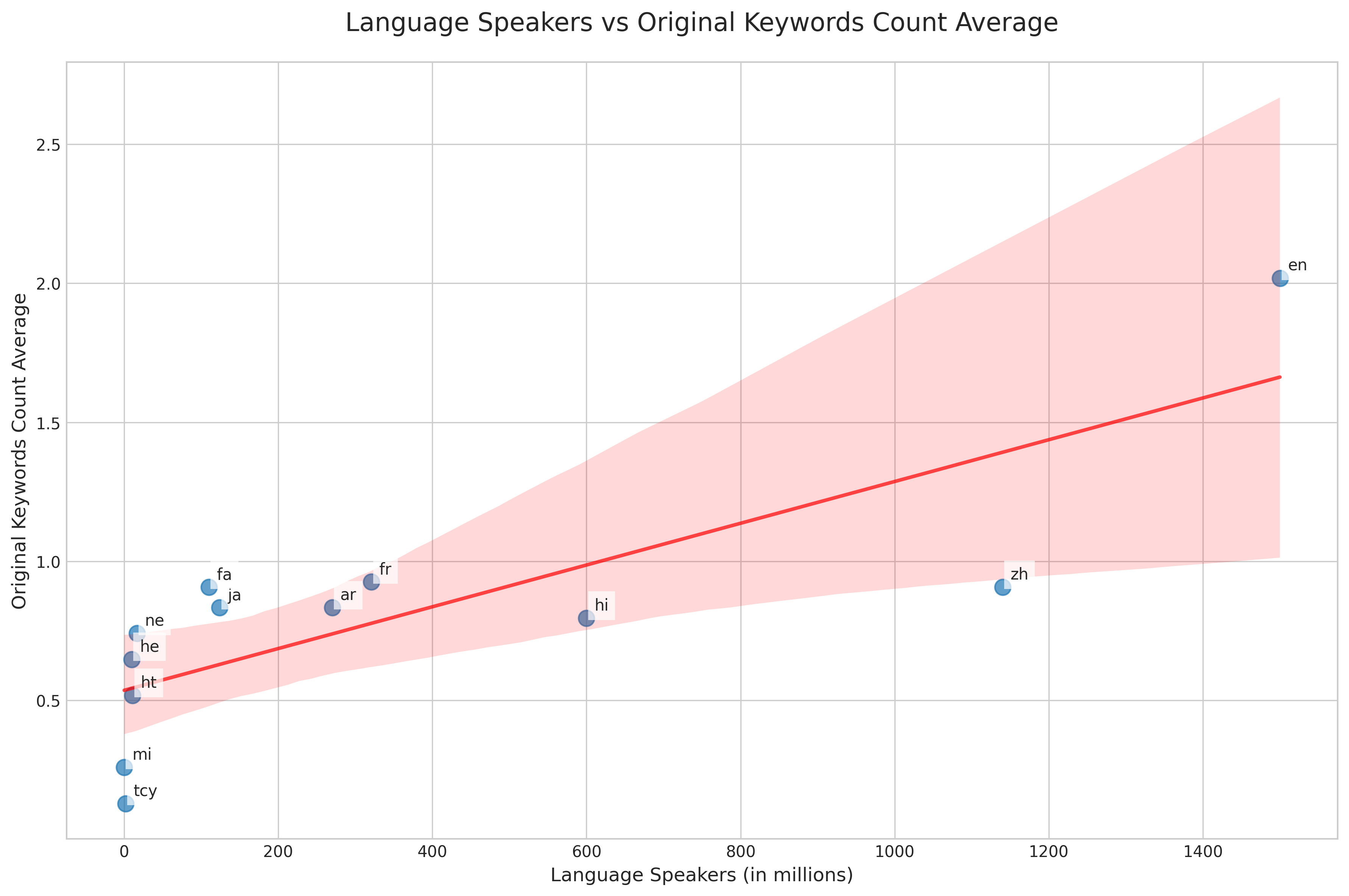}
    \caption{Relationship between \# of Speakers and Target Words for LLaMA-70B}
    \label{fig:keywords_vs_speakers}
\end{figure}

The correlations we find are more striking when visualized, and reveal nuances to the data that the aggregated figures alone do not provide. Figure~\ref{fig:keywords_vs_speakers} shows the relationship between the number of speakers of a language and the average count of target words detected in LLaMA-70B responses. The trend indicates that languages with more speakers tend to have higher target word coverage, suggesting that these languages are better represented in the training data, leading to more accurate responses. The variance in languages with very few speakers probably exists because the online data for these languages, both in terms of quality and quantity, is likely to vary wildly.

\begin{figure}[htbp]
    \centering
    \includegraphics[width=0.6\textwidth]{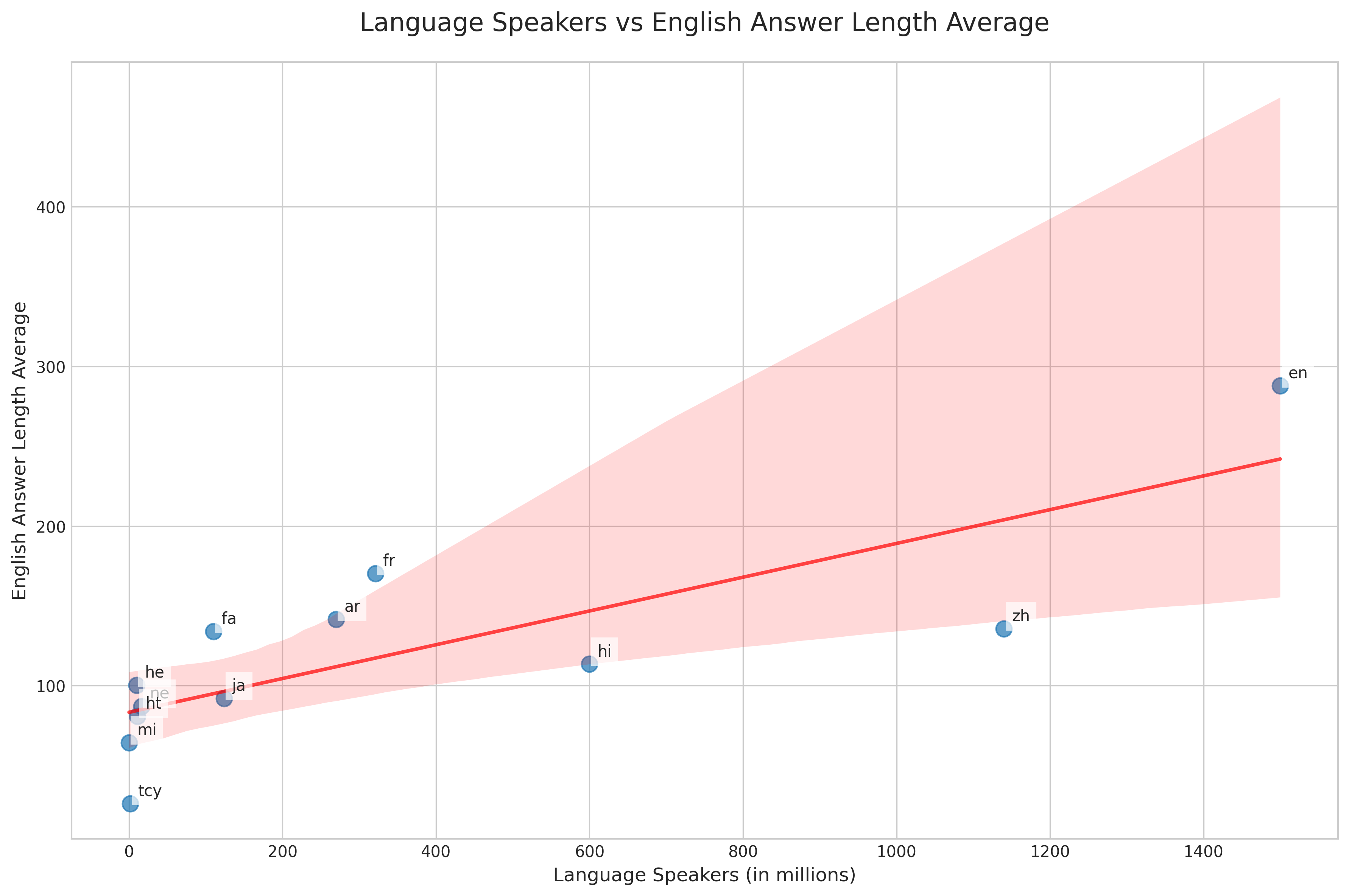}
    \caption{Relationship between \# of Speakers and Response Length (when translated to English) for LLaMA-70B}
    \label{fig:answer_length_vs_speakers}
\end{figure}

Figure~\ref{fig:answer_length_vs_speakers} illustrates the relationship between the number of speakers and the average response length (translated to English) for LLaMA-70B. The correlation suggests that languages with fewer speakers tend to have shorter responses, which may indicate terseness or under representation in training data. Interestingly, we did not observe the phenomenon for the smaller model, which was terser in general.

For some languages, counting language speakers entails entanglement in complicated social and political issues. Many of these language figures are disputed, which may also contribute to the noisiness of our data. When language figures differed wildly across sources, we averaged them.

\begin{figure}[htbp]
    \centering
    \includegraphics[width=0.8\linewidth]{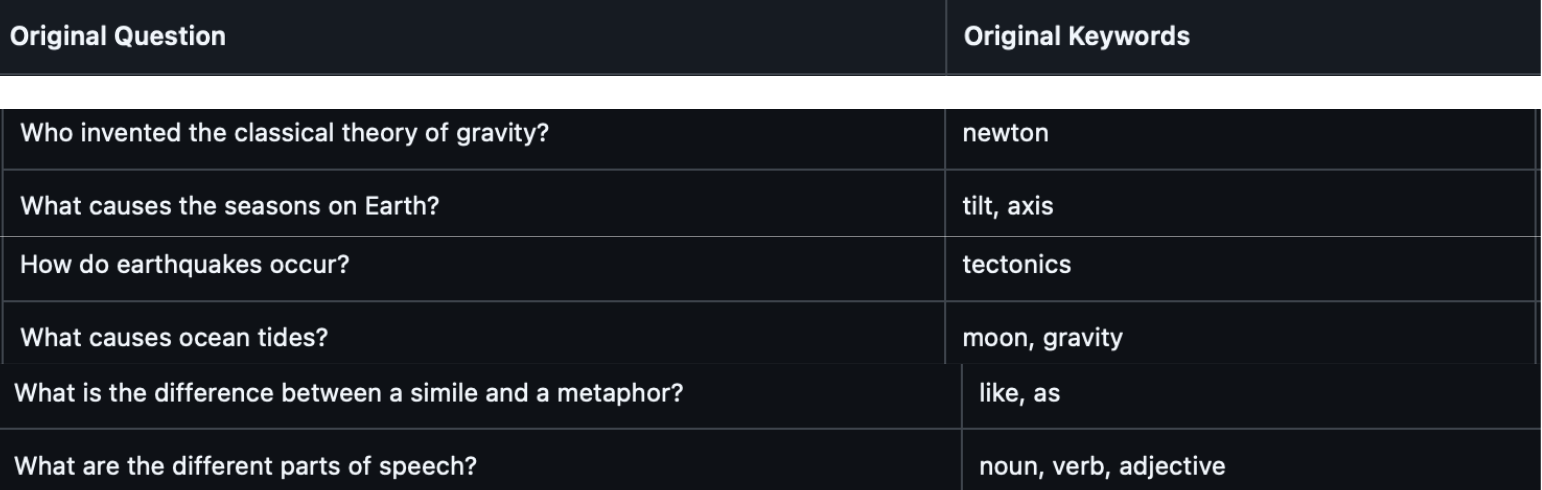}
    \caption{Sample questions from our dataset (in English) and their corresponding keywords}
    \label{fig:sample_questions}
\end{figure}

In Figure \ref{fig:sample_questions} we present sample original questions from our dataset (in English) and their corresponding keywords that the raters were asked to look for and mark in the translated LLM responses. The questions and the keywords were hand-curated and chosen from a breadth of topics relevant to a high/middle school curriculum.

\begin{figure}
    \centering
    \includegraphics[width=0.75\linewidth]{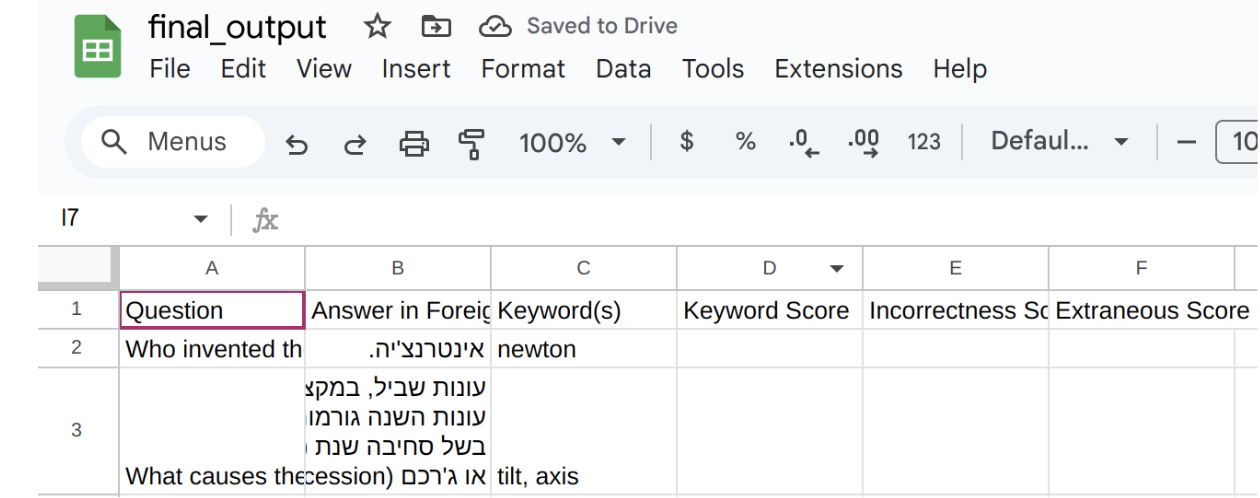}
    \caption{Sample sheet given to the manual evaluator to rate non-English responses}
    \label{sample_evaluation_sheet}
\end{figure}

Figure \ref{sample_evaluation_sheet} presents a screenshot from an evaluation sheet that were given to the manual evaluators. This particular example is for a Hebrew evaluator. For keyword matching, manual evaluators were asked to look for hand-curated keywords in the translated answers and assign "Keyword", “Incorrectness” and “Extraneous” scores accordingly.

The sample prompt given to the LLM to get its responses in non-English languages was: 
\begin{center}
    \textit{You are a helpful assistant. Please answer the question in the same language that it is asked in. $<$QUESTION$>$}
\end{center}

\end{document}